\documentclass{article}

\usepackage[numbers]{natbib}

     \usepackage[final]{neurips_2020}

\usepackage[utf8]{inputenc} 
\usepackage[T1]{fontenc}    
\usepackage{url}            
\usepackage{booktabs}       
\usepackage{amsfonts}       
\usepackage{nicefrac}       
\usepackage{microtype}      
\usepackage{algorithm}
\usepackage{algorithmic}
\usepackage{wrapfig}
\usepackage{rotating}

\usepackage{graphicx}
\usepackage{amsmath}
\usepackage{subfigure}
\usepackage{booktabs} 
\usepackage[pagebackref=true,breaklinks=true,letterpaper=true,colorlinks,bookmarks=false,citecolor=blue]{hyperref}

\usepackage[utf8]{inputenc} 
\usepackage[T1]{fontenc}    
\usepackage{url}            
\usepackage{booktabs}       
\usepackage{amsfonts}       
\usepackage{nicefrac}       
\usepackage{microtype}      

\usepackage{graphicx}
\usepackage{subfigure}
\usepackage{amsmath}
\usepackage{amssymb}
\usepackage{caption}
\usepackage{courier}
\usepackage{dsfont}
\usepackage{mathtools}
\usepackage{dsfont}
\usepackage{multirow}
\usepackage{tabularx}
\usepackage{arydshln}
\usepackage{pifont}
\usepackage{algorithm}
\usepackage{algorithmic}
\usepackage{todonotes}
\usepackage{setspace}
\usepackage{wrapfig}
\usepackage{lipsum}

\usepackage{enumitem}
\setlist[itemize]{leftmargin=5mm, nolistsep}

\usepackage{color}
\usepackage{xcolor}

\newcommand{\R}{\mathbb{R}}

\newcommand{\bN}{\textbf{N}}
\newcommand{\calT}{\mathcal{T}}
\newcommand{\calL}{\mathcal{L}}
\newcommand{\Lin}{\mathcal{L}^\mathrm{inner}}

\newcommand{\Lmeta}{\mathcal{L}^\mathrm{meta}}

\usepackage{todonotes}
\usepackage{xcolor}

\title{Meta-Neighborhoods}

\author{%
  Siyuan Shan \\
  Department of Computer Science\\
  University of North Carolina at Chapel Hill\\
  \texttt{siyuanshan@cs.unc.edu} 
  
  \And
  Yang Li \\
  Department of Computer Science\\
  University of North Carolina at Chapel Hill\\
  \texttt{yangli95@cs.unc.edu} 
  
  \And
  Junier B. Oliva \\
  Department of Computer Science\\
  University of North Carolina at Chapel Hill\\
  \texttt{joliva@cs.unc.edu} 
}
\DeclareUnicodeCharacter{2212}{-}
\begin{document}
\setlength{\columnsep}{10pt}
\setlength{\intextsep}{10pt}

\maketitle

\begin{abstract}
Making an adaptive prediction based on one's input is an important ability for general artificial intelligence. In this work, we step forward in this direction and propose a semi-parametric method, Meta-Neighborhoods, where predictions are made adaptively to the neighborhood of the input. We show that Meta-Neighborhoods is a generalization of $k$-nearest-neighbors. Due to the simpler manifold structure around a local neighborhood, Meta-Neighborhoods represent the predictive distribution $p(y \mid x)$ more accurately. To reduce memory and computation overhead, we propose induced neighborhoods that summarize the training data into a much smaller dictionary. A meta-learning based training mechanism is then exploited to jointly learn the induced neighborhoods and the model. Extensive studies demonstrate the superiority of our method\footnote{The code is available at \url{https://github.com/lupalab/Meta-Neighborhoods}}.%
\end{abstract}

\section{Introduction}
\begin{wrapfigure}{R}{0.3\textwidth}
\centering
\vspace{-10pt}
\includegraphics[width=0.3\textwidth]{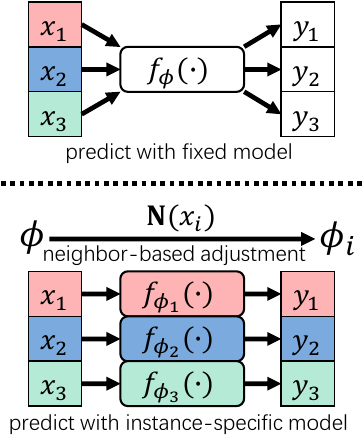} 
\caption{Top: traditional parametric models. Bottom: our per-instance adapted model.}
\vspace{-15pt}
\label{fig:simple_framework}
\end{wrapfigure}
Discriminative machine learning models typically learn the predictive distribution $p(y \mid x)$. There are two paradigms to build a model, parametric methods and non-parametric methods \cite{hastie2009elements}. Parametric methods assume that a set of \emph{fixed} parameters $\theta$ dominates the predictive distribution, i.e., $p(y \mid x ; \theta)$. The training process estimates $\theta$ and then discard the training data completely, as the learned parameters $\theta$ are responsible for the following prediction. This paradigm has proven effective, however, it leaves the entire burden on learning a complex predictive distribution over potentially large support. Non-parametric models differ in that the number of parameters scales with data. They typically reuse the training data during the testing phase to make predictions. For instance, the well-known $k$-nearest neighbor (KNN) estimator often achieves surprisingly good results by leveraging neighbors from the training data, which reduces the problem to a much simpler local-manifold. Despite its flexibility, non-parametric methods are required to store the training data and traverse them during testing, which may impose significant memory and computation overhead for large training sets.

In this work, we combine the merits of both paradigms and propose a semi-parametric method called Meta-Neighborhoods. The main body of Meta-Neighborhoods is a parametric neural network, but we adapt its parameters to a local neighborhood in a non-parametric scheme. The prediction is made on the local manifold by the adapted model. Fig. \ref{fig:simple_framework} illustrates the difference between traditional parametric models and the proposed model. Inspired by the success of inducing point methods from sparse Gaussian process literature \cite{snelson2006sparse,titsias2009variational} to alleviate the storage burden and reduce time complexity, we learn induced neighborhoods, which summarize the training data into a much smaller dictionary. The induced neighborhoods and the neural network parameters are learned jointly.

Our model is also closely related to locally linear embeddings \cite{roweis2000nonlinear}, which reconstructs the non-linear manifold with locally linear approximation around each neighborhood. In our method, we adapt an initial model (not necessarily linear) to local neighborhoods. Since the local manifold is much simpler, we expect the adapted model can better capture the predictive distribution. Overall, it learns a better discriminative model on the entire support.

Our method imposes challenges of adapting the initial model since the local neighborhoods usually do not contain enough training instances to independently adapt the model, and the induced neighborhoods contain even fewer instances. Inspired by the few-shot and meta-learning literature \cite{finn2017model}, we propose a meta-learning based training mechanism, where we learn an initial model so that it adapts to the local neighborhood after only several finetuning steps over a few instances inside the neighborhood.

The prediction process of our model remains flexible by following a non-parametric scheme. An input $x$ is first paired with its neighbors by querying the induced dictionary. The initial model is adapted to its neighborhood by finetuning several steps on the neighbors. We then predict the target $y$ using the adapted model.

Our contributions are as follows:
\begin{itemize}
    \item We combine parametric and non-parametric methods in a meta-learning framework.
    \item We propose Meta-Neighborhoods to jointly learn the induced neighborhoods and an adaptive initial model, which can adapt its parameters to a local neighborhood according to the input through both finetuning and a proposed instance-wise modulation scheme, iFiLM.
    \item  Extensive empirical studies demonstrate the superiority of Meta-Neighborhoods for both regression and classification tasks.
    \item We empirically find the induced neighbors are semantically meaningful: they represent informative boundary cases on both realistic and toy datasets; they also capture sub-category concepts even though such information is not given during training.
\end{itemize}

\section{Method}
\paragraph{Problem Formulation}
Given a training set $\mathcal{D}=\{(x_i,y_i)\}_{i=1}^{N}$ with $N$ input-target pairs, we learn a discriminative model $f_\phi(x)$ and a dictionary $M=\{(k_{j},v_{j})\}_{j=1}^S$ jointly from $\mathcal{D}$. The learned dictionary stores the neighbors induced from the training set, where $S$ is the number of induced neighbors. Just like the real training set $\mathcal{D}$, the dictionary stores input-target pairs as key-value pairs ($k_j$,$v_j$), where both the keys and the values are learned end-to-end with the model parameters $\phi$ in a meta-learning framework. For classification tasks, $v_j$ is a vector representing the class probabilities while for regression tasks $v_j$ is the regression target. In the following text, we will use the terms ``induced neighbors" and ``learnable neighbors" interchangeably. We defer the exact training mechanism to Section \ref{sec:joint_meta_learning}.


\subsection{Predict with Induced Neighborhoods}

\label{sec:meta-neighbor}
In this section, we assume access to the learned neighborhoods in $M$ and the learned model $f_\phi$. Different from the conventional parametric setting, where the learned model is employed directly to predict the target, we adapt the model according to the neighborhoods retrieved from $M$ and the adapted model is the one responsible for making predictions. Specifically, for a test data $x_i$, relevant entries in $M$ are retrieved in a soft-attention manner by comparing $x_i$ to $k_j$ via a attention function $\omega$. The retrieved entries are then utilized to finetune $f_\phi$ following
\begin{equation}
    \phi_i \leftarrow \phi - \alpha \nabla_\phi \sum_{j=1}^{S}\omega(x_i, k_j) L(f_\phi(k_j),v_j),
    \label{eq:inner}
\end{equation}
where $\alpha$ is the finetuning step size. 
Note we weight the loss terms of all dictionary entries by their similarities to the input test data $x_i$. The intuition is that nearby neighbors play a more important role in placing $x_i$ into the correct local manifold.
Since the model $f_\phi$ is specially trained in a meta-learning scheme, it adapts to the local neighborhoods after only a few finetuning steps.

To better understand our method, we draw connections to other well-known techniques. The above prediction process is similar to a one-step EM algorithm. Specifically, the dictionary querying step is an analogy to the Expectation step, which determines the latent variables (in our case, the neighborhood assignment). And the finetuning step is similar to the Maximization step, which maximizes the expected log-likelihood. We can also view this process from a Bayesian perspective, where the initial parameter $\phi$ is an empirical prior estimated from data; posteriors are derived from neighbors following the finetuning steps. The predictive distribution with posterior is used for the final predictions.

\begin{figure*}[ttt!]
\begin{tabular}{cc}
\begin{minipage}{.43\textwidth}
    \centering
    \includegraphics[width=\textwidth]{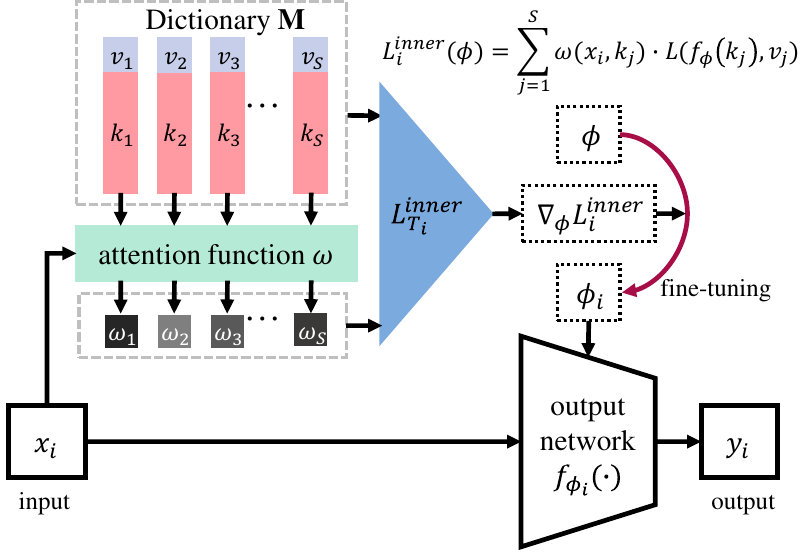}
    \captionof{figure}{
        Model overview.
    } \label{fig:model}
\end{minipage} &
\hfill
\scalebox{0.7}{
    \begin{minipage}{.7\textwidth}
    \begin{algorithm}[H]
    \captionof{algorithm}{\textsc{Meta-Neighborhoods: Training Phase}} \label{alg:train}
        \begin{spacing}{1.1}
        \begin{algorithmic}[1]
          \REQUIRE $\omega$: similarity metric, $\eta$: outer loop learning rate
\STATE Initialize $\theta,\phi,\alpha,M=\{(k_{j},v_{j})\}_{j=1}^S$
\WHILE{not done}
\STATE Sample a batch of training data $\{(x_{i},y_{i})\}_{i=1}^B$
  \FORALL{$(x_i,y_i)$ in current batch}
 \STATE Compute the feature vector $z_{i} = \mu_{\theta}(x_{i})$
 \STATE Compute $\Lin_{i}(\phi)=\sum_{j=1}^{S}\omega(z_i, k_j) L(f_\phi(k_j),v_j)$
 \STATE Finetune $\phi$:
 $\phi_{i}=\phi-\alpha \nabla_{\phi} \Lin_{i}(\phi)$
 \ENDFOR
 \STATE Compute $\Lmeta(\phi,\theta,M,\alpha)=\frac{1}{B} \sum_{i=1}^{B} L(f_{\phi_i}(\mu_\theta(x_i)),y_i)$
 \STATE Update model parameters 	$\Theta = \{\theta,\phi,M,\alpha\}$ \\
 using gradient descent as $\Theta \leftarrow \Theta - \eta \nabla_\Theta \Lmeta$
\ENDWHILE
        \end{algorithmic}
    \end{spacing}
    \end{algorithm}
    \end{minipage}
}
\end{tabular}
\end{figure*}

\subsection{Joint Meta Learning}
\label{sec:joint_meta_learning}
Above, we assume access to given $M$ and $f_\phi$, in this section, we describe our meta-learning mechanism to train them jointly.
The training strategy of Meta-Neighborhoods resembles MAML \cite{finn2017model} in that both adopt a nested optimization procedure, which involves an inner loop update and an outer loop update in each iteration. Note that in contrast to MAML, we are solving a general discriminative task rather than a few-shot task. Given a batch of training data $\{x_i,y_i\}_{i=1}^{B}$ with a batch size $B$, in the inner loop we finetune the initial parameter $\phi$ to $\phi_i$ in a similar fashion to \eqref{eq:inner}. With $\phi$ individually finetuned for each training data $x_i$ using its corresponding neighborhoods, we then jointly train the model parameter $\phi$, the dictionary $M$ as well as the inner loop learning rate $\alpha$ in the outer loop using the following meta-objective function
\begin{equation}
\begin{aligned}
\label{eq:outer-loss}
  \Lmeta(\phi,M,\alpha)=\frac{1}{B}\sum_{i=1}^{B}L(f_{\phi_i}(x_i),y_i) 
    =\frac{1}{B}\sum_{i=1}^{B}L(f_{\phi-\alpha \nabla_\phi \Lin_{i}}(x_i),y_i),
\end{aligned}
\end{equation}
where $\Lin_{i}(\phi)=\sum_{j=1}^{S}\omega(x_i, k_j) L(f_\phi(k_j),v_j)$ according to \eqref{eq:inner}. We set $\alpha$ to be a learnable scalar or diagonal matrix. $\Lmeta$ encourages learning shared $\phi$, $M$, and $\alpha$ that are widely applicable for data with the same distribution as the training data. An overview of our model is shown in Fig.~\ref{fig:model}.

Parameter $\phi$ serves as initial weights that can be quickly adapted to a specified neighborhood.
This meta training scheme effectively tackles the overfitting problem caused by the limited number of finetuning instances, as it explicitly optimizes the generalization performance after finetuning. 

For high-dimensional inputs such as images, learning $k_j$ in the input space could be prohibitive. Therefore, we employ a feature extractor $\mu_\theta$ to extract the feature embedding $z_{i}=\mu_\theta(x_i)$ for each $x_i$ and learn $k_j$ in the embedding space. We accordingly modify \eqref{eq:inner} to
\begin{equation}
    \phi_i \leftarrow \phi - \alpha \nabla_\phi \sum_{j=1}^{S}\omega(\mu_\theta(x_i), k_j) L(f_\phi(k_j),v_j),
    \label{inner_loop_feature_space}
\end{equation}
where the attention function $\omega$ is employed in embedding space.
The meta-objective is accordingly modified as $\Lmeta(\phi,\theta,M,\alpha) = \frac{1}{B}\sum_{i=1}^{B}L(f_{\phi_i}(\mu_\theta(x_i)),y_i)$.
We train $\theta$ and other learnable parameters jointly. Note that the model without a feature extractor can be viewed as a special case where $\mu_\theta$ is an identity mapping. 
The pseudocode of our training algorithm is given in Algorithm \ref{alg:train}. 

It is also desirable to adjust $\mu_{\theta}$ per-instance. However, when $\mu_\theta$ is a deep convolution neural network, tuning the entire feature extractor $\mu_{\theta}$ is computationally expensive. Inspired by FiLM \cite{perez2018film}, we propose instance-wise FiLM (iFiLM) that adjusts the batch normalization layers individually for each instance.
Suppose $a^l \in \R^{B\times C^l\times W^l\times H^l}$ is the output of the $l_{th}$ Batch Normalization layer $\textbf{BN}^{l}$, where $B$ is batch size, $C^l$ is the number of channels, $W^l$ and $H^l$ are the feature map width and height. Along with each $\textbf{BN}^{l}$, we define a learnable dictionary $M^{l}=\{k^l_j,\gamma^l_j, \beta^l_j\}_{j=1}^{S_l}$ of size $S^l$. $k^l_j$ are the keys used for querying. $\gamma^l_j,\beta^l_j \in \R^{C^l}$ represent the scale and shift parameters used for adaptation respectively. 
When querying $M^{l}$, the outputs $a^l$ are first aggregated across their spatial dimensions using global pooling, i.e.
$
    g^l=\textbf{GlobalAvgPool}(a^l)\in \R^{B\times C^l}.
$
Then, the instance-wise adaptation parameters $\widehat{\gamma^{l}_i}$ and $\widehat{\beta^{l}_i}$ are computed as
\begin{equation}
\label{eq:ifilm_attention}
    \widehat{\gamma^{l}_i}=\sum_{j=1}^{S^l}\omega(g^l_i,k^l_j)\gamma^l_j \in \R^{C^l} \quad\quad
    \widehat{\beta^{l}_i}=\sum_{j=1}^{S^l}\omega(g^l_i,k^l_j)\beta^l_j \in \R^{C^l},
\end{equation}
where $\omega$ is defined as in \eqref{eq:inner} and $i\in\{1,2,\ldots,B\}$. Following FiLM \cite{perez2018film}, each individual activation $a^l_i$ is then adapted with an affine transformation
$
    \widehat{\gamma^{l}_i} \cdot a^l_i + \widehat{\beta^{l}_i}.
$
Note the transformation is agnostic to spatial position, which helps to reduce the number of learnable parameters in the dictionary $M$.


%

\subsection{Other Details and Considerations}
\label{details_considerations}
In this section, we discuss further implementation details. We also motivate our method from the perspective of $k$-nearest neighbor (KNN).

\paragraph{Similarity Metrics} 
To implement the attention function $\omega$ in \eqref{eq:inner}\eqref{inner_loop_feature_space}\eqref{eq:ifilm_attention}, we need a similarity metric to compare a input $x_i$ with each key $k_j$. We try two types of metrics, cosine similarity and negative Euclidean distance.
The similarities of $x_i$ to all keys are normalized using a softmax function with a temperature parameter $T$ \cite{hinton2015distilling}, i.e.,
\begin{equation}\label{eq:sim}
    \omega(x_i, k_j) = \frac{\exp(\text{sim}(x_i, k_j)/T)}{\sum_{s=1}^{S}\exp(\text{sim}(x_i, k_s)/T)},
\end{equation}
where $\text{sim}(\cdot)$ represents the similarity metric.

\paragraph{Initialization of the Dictionary}
Since we use similarity-based attention function $\omega$ in \eqref{inner_loop_feature_space}, we would like to initialize the key $k_j$ to have a similar distribution to $z_i=\mu_\theta(x_i)$, otherwise, $k_j$ cannot receive useful training signal at early training steps. To simplify the initialization, we follow \cite{gidaris2018dynamic} to remove the non-linear function (e.g. ReLU) at the end of $\mu_\theta$ so that features extracted by $\mu_{\theta}$ are approximately Gaussian distributed. With this modification, we can simply initialize $k_j$ with Gaussian distribution. 

\paragraph{Cosine-similarity Based Classification}
Since the model $f_\phi$ is finetuned using the learned dictionary in the inner loop, the quality of the dictionary has a significant impact on final performance. Typical neural network classifiers employ dot product to compute the logits, where the magnitude and direction could both affect the results. Therefore, the model needs to learn both the magnitude and the direction of $k_j$. To alleviate the training difficulty, we eliminate the influence of magnitude by using a cosine similarity classifier, where only the direction of $k_j$ can affect the computation of logits. Cosine similarity classifiers have been adopted to replace dot product classifiers for few-shot learning \cite{gidaris2018dynamic,chen2018a}
 and robust classification \cite{yang2018robust}.

\paragraph{Relationship to KNN}
Below, we show that Meta-Neighborhoods can be derived as a direct generalization of KNN under a multi-task learning framework. Considering a regression task where the regression target is a scalar, the standard view of KNN is as follows. First, aggregate the $k$-nearest neighbors of a query $\tilde{x_i}$ from the training set $\mathcal{D}$ as $\bN(\tilde{x_i})=\{(x_j, y_j)\}_{j=1}^k \subset \mathcal{D}$.
Then, predict an  average of the responses in the neighborhood: $\hat{y} = \frac{1}{k} \sum_{j=1}^k y_j$. 

Instead of simply performing an average of responses in a neighborhood, we frame KNN as a solution to a multi-task learning problem with tasks corresponding to individual neighborhoods as follows.
Here, we take each query (test) data $\tilde{x_i}$ as a single task, $\calT_{i}$. To find the optimal estimator on the neighborhood $\bN(\tilde{x_i})=\{(x_j, y_j)\}_{j=1}^k$, we optimize the following loss 
$\calL_{\calT_{i}}(f_{i})=\frac{1}{k}\sum_{j=1}^{k}L(f_i(x_{j}),y_{j})$ where $L$ is a supervised loss, and $f_i$ is the estimator to be optimized.
For example, for MSE-based regression the loss for each task is $\mathcal{L}_{\mathcal{T}_{i}}(f_i)= \frac{1}{k}\sum_{j=1}^{k}(f_i(x_{j})-y_{j})^2$. If one takes $f_i$ to be a constant function $f_i(\eta_{j}) = C_i$, then the loss is simply $\mathcal{L}_{\mathcal{T}_{i}}(f_i) =\frac{1}{k}\sum_{j=1}^{k}(C_i-\zeta_{j})^{2}$, which leads to an optimal $f_i(\tilde{x_i}) = C_i = \frac{1}{k} \sum_{j=1}^k \zeta_j$, the same solution as traditional KNN. Similar observations hold for classification.
\emph{Thus, given neighborhood assignments, one can view KNN as solving for individual tasks in the special case of a constant estimator $f_i(x_{j})= C_i$}.

With the multi-task formulation of KNN, we can generalize KNN to derive our Meta-Neighborhoods method by considering a non-constant estimator as $f_i$.
For instance, one may take $f_i$ as a parametric output function $f_{\phi_i}$ (e.g. a linear model or neural networks), and finetune the parameter $\phi$ to $\phi_i$ for a data $x_i$ according to the loss on neighborhood $\textbf{N}(x_i)$.
Instead of fitting a single label on the neighborhood, a parametric approach attempts to fit a richer (e.g. linear) dependency between input features and labels for the neighborhood. In addition, the multi-task formulation gives rise to a way of constructing meta-learning episodes. Also, we learn both neighborhoods and the function $f_\phi$ jointly in our Meta-Neighborhoods framework.

\section{Related Work}
\paragraph{Memory-augmented Neural Networks}
Augmenting neural network with memory has been studied in the sentinel work Neural Turing Machine \cite{graves2014neural}, where a neural network can read and write an external memory to record and change its state. 
Recent works that utilize the memory modules generally fall into two categories. One category modifies the memory modules according to hand-crafted rules. For instance, previous works tackling few-shot classifications add a new slot to the memory when the label of a given data does not match the labels of its $k$-nearest neighbors from the memory \cite{cai2018memory} or the given data is misclassified \cite{ramalho2019adaptive}. \cite{sprechmann2018memory} adopts a fixed-size memory that acts as a circular buffer for life-long learning. Another category uses a fully-differentiable memory module and trains it together with neural networks by gradient descent. This type of memory has been explored for knowledge-based reasoning \cite{graves2016hybrid}, sequential prediction \cite{sukhbaatar2015end} and few-shot learning \cite{kaiser2017learning,santoro2016meta}. Our work also utilizes a differentiable memory but is used to capture local manifold and improve the general discriminative learning performance. 

\paragraph{Meta-Learning}
Representative meta-learning algorithms can be roughly categorized into two classes: initialization based and metric-learning based. Initialization based methods, such as MAML \cite{finn2017model}, learn a good initialization for model parameters so that several gradient steps using a limited number of labeled examples can adapt the model to make predictions for new tasks. To further improve flexibility, Meta-SGD \cite{li2017meta} learns coordinate-wise inner learning rates, and curvature information is considered in \cite{park2019meta} to transform the gradients in the inner optimization. Metric-learning based methods focus on using a distance metric on the feature space to compare query set samples with labeled support set samples. Examples include cosine similarity \cite{vinyals2016matching} or Euclidean distance \cite{snell2017prototypical} to support examples. A learned relation module is employed in  \cite{sung2018learning}.

Our model is in a similar vein to the initialization based method: each test sample can be regarded as a new task, and we meta-learn a dictionary that adapts the initial model to a local neighborhood by finetuning over queried neighbors.
A recent work Meta AuXiliary Learning (MAXL) also explores meta-learning techniques to improve classification performance, where a label generator is meta learned to generate auxiliary labels so that the auxiliary task trained together with the primary classification task can improve the primary performance.

\begin{figure}[t!]
\center
\subfigure[Iteration 0]{\includegraphics[width=0.3\linewidth]{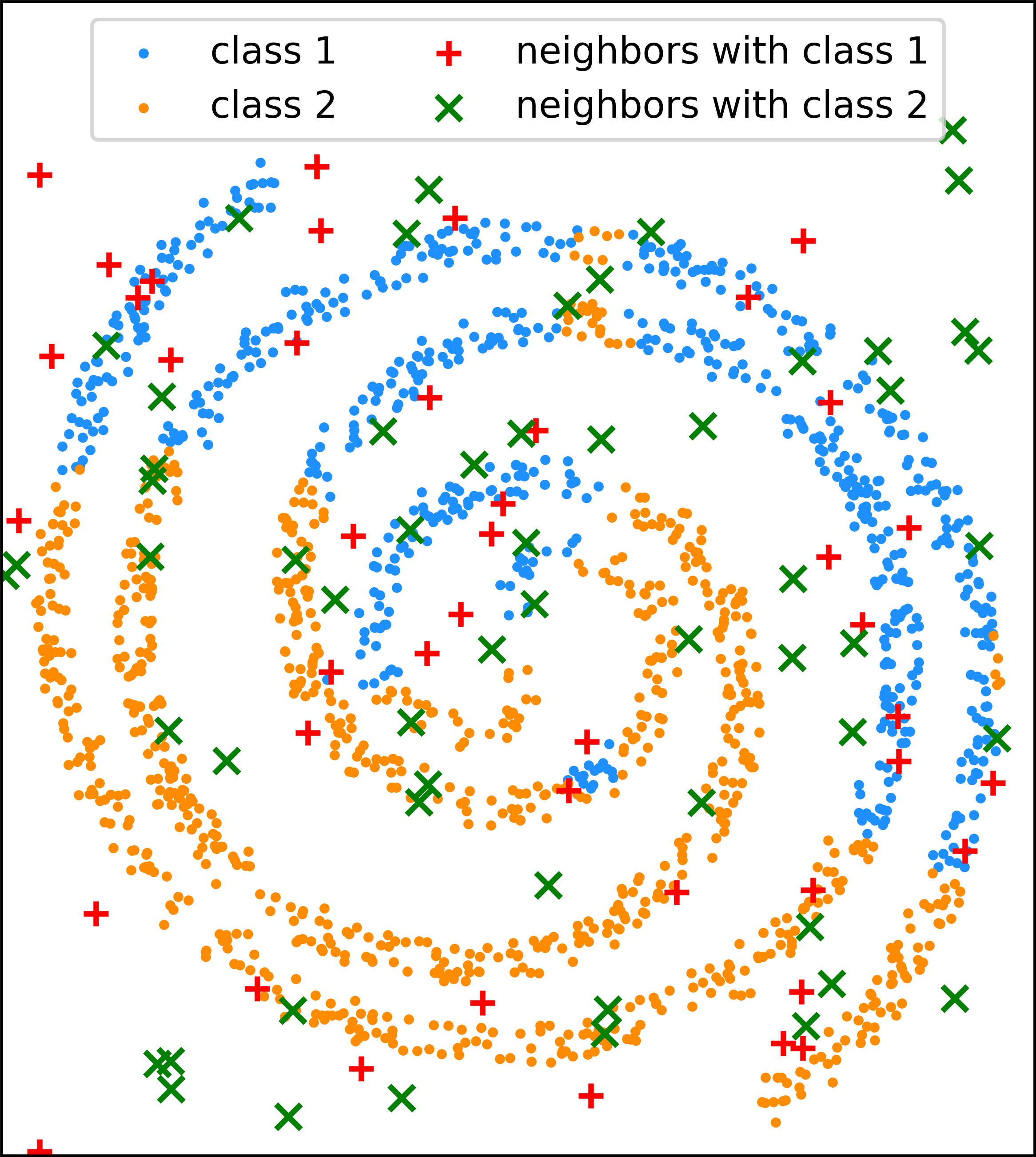}}
\subfigure[Iteration 800]{\includegraphics[width=0.3\linewidth]{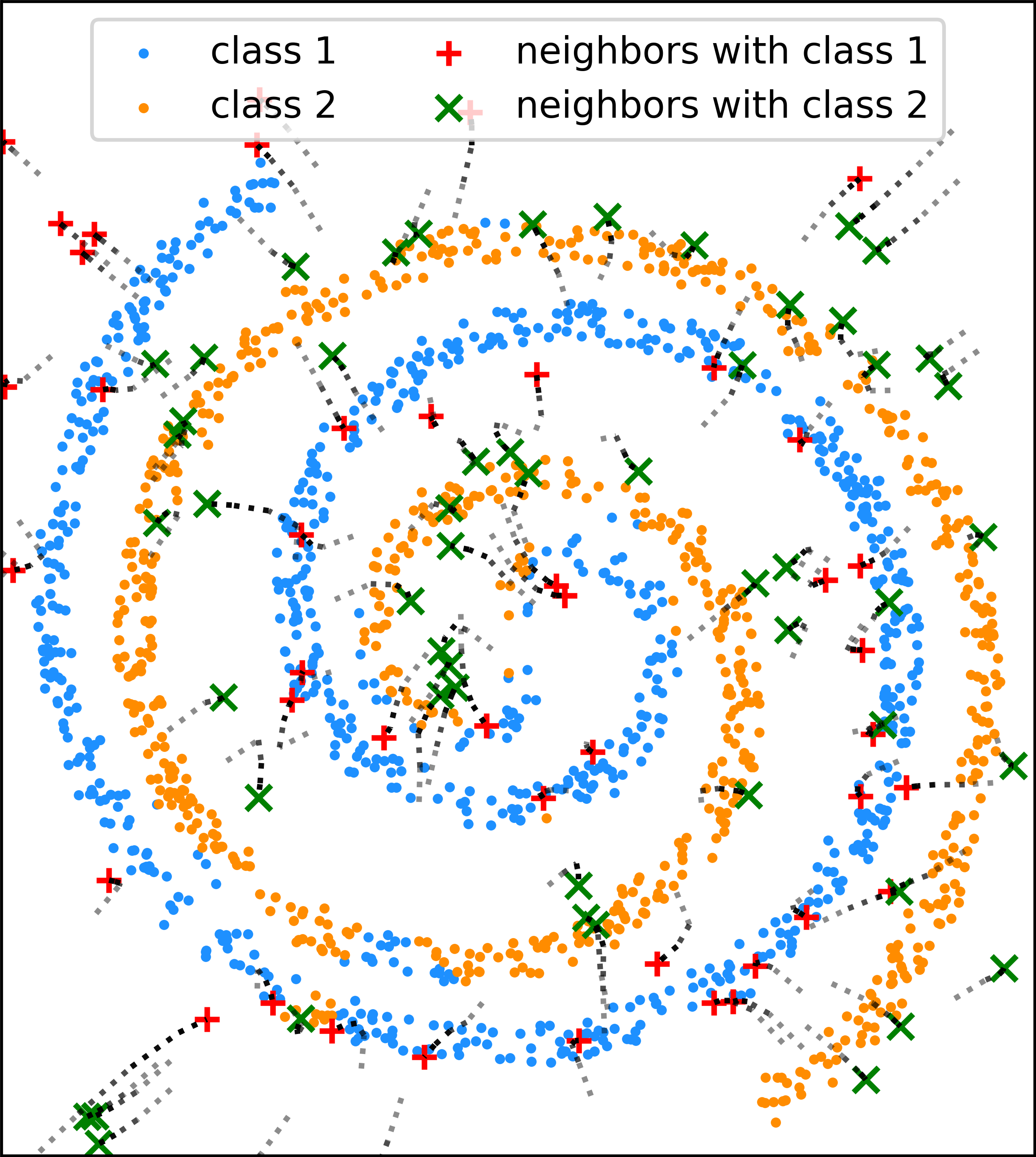}}
\subfigure[Iteration 6400]{\includegraphics[width=0.3\linewidth]{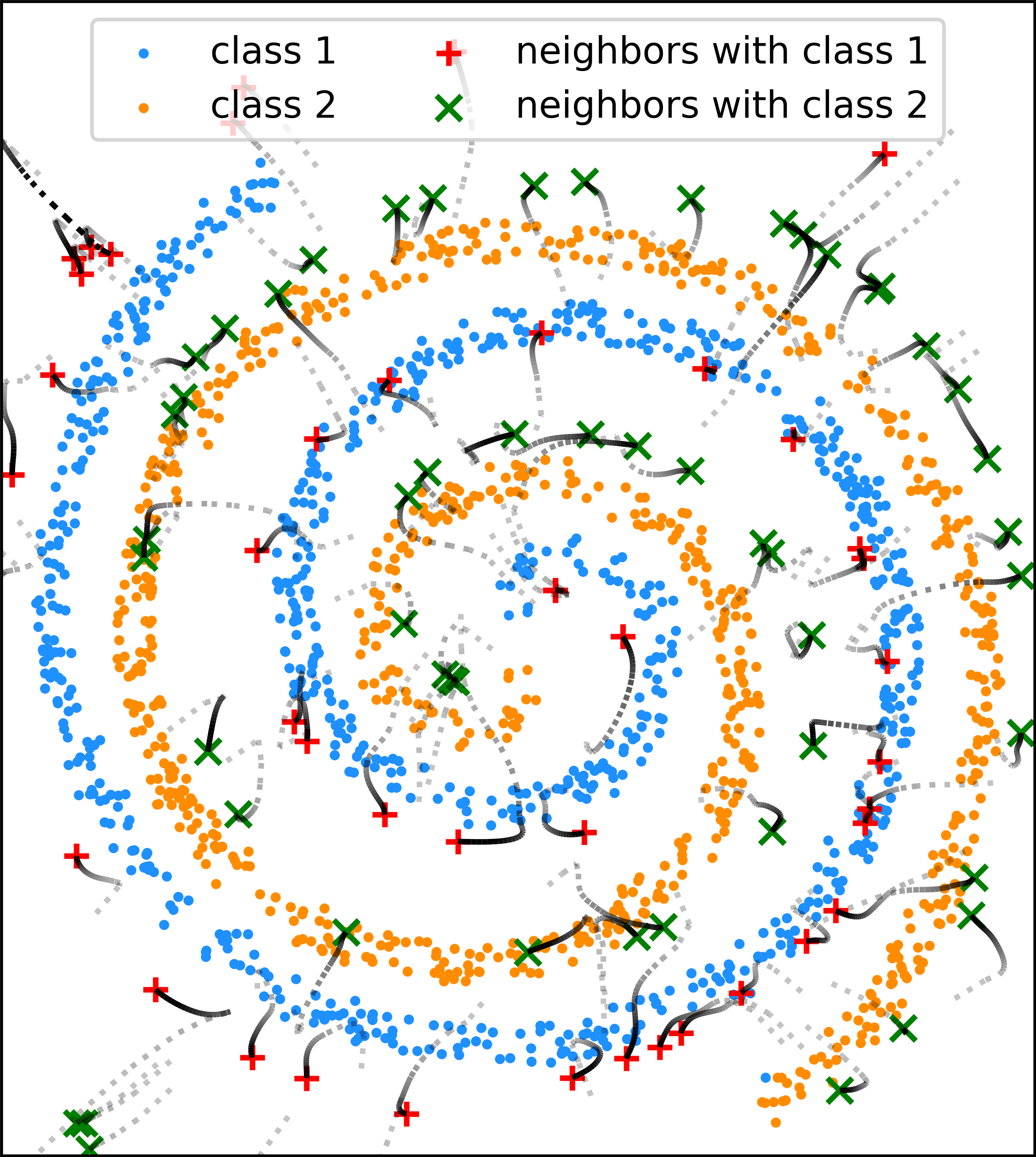}}
  \caption{Evolution of learnable neighbors and classification results on the test data during training. Two classes are two spirals. Binary predictions for the test set are shown as blue and yellow points. Learnable neighbors are first randomly initialized in (a), then optimized in (b) (c). The dotted black lines are the trajectories of learnable neighbors through training iterations. A video of the optimization process is given in the supplemental materials.}
  \label{fig1}
\end{figure}

\section{Experiments}
In this section, we conduct experiments for both classification and regression tasks. To demonstrate the benefits of making predictions in local neighborhoods, we compare to the $\textit{vanilla}$ model where the same network architecture is used but without the learnable neighborhoods. We also compare to MAXL \cite{liu2019self} for classification task.

\subsection{Toy Example: Binary Classification of the Concentric Spiral Dataset}
To investigate the behavior of the induced neighbors and how they assist the parametric model to make predictions, we first classify on a 2D toy dataset.
In this binary classification task, points from two classes are placed in concentric spirals with a non-linear decision boundary. Although a linear classifier is incapable of capturing this decision boundary, we show that tuning a linear classifier with induced neighbors gives an overall non-linear classifier. In addition, our learned neighborhoods capture the critical manifold structure and concentrate at boundary cases; we also observe semantically relevant learned neighbors in higher dimensions (see Appendix \ref{vis_neighbor_sub_category} and \ref{sub_category_discovery})

In Fig.~\ref{fig1}, we visualize the evolution of the learned neighbors and the decision boundary. 
The learnable neighbors (shown in green and red markers for two classes respectively) are first initialized with random keys and values. As we train the model and the neighbors, the learned neighbors are gradually driven to important manifold locations as in Fig. \ref{fig1} (b) and (c). After training, the linear classifier adapted to local neighborhoods can accurately classify test examples. We use 100 induced neighbors. The labels of neighbors (values in the dictionary) are fixed after random initialization for illustration purposes. The 2D locations of neighbors (keys in the dictionary) are updated with the model. We use negative Euclidean distance as the similarity metric in \eqref{eq:sim} and set $T$ to 0.1. 

\subsection{Image Classification}
In this section, we evaluate 9 datasets with different complexities and sizes: MNIST \cite{lecun1998gradient}, MNIST-M\cite{ganin2016domain}, PACS\cite{li2017deeper}, SVHN \cite{goodfellow2013multi}, CIFAR-10 \cite{krizhevsky2009learning}, CIFAR-100 \cite{krizhevsky2009learning}, CINIC-10 \cite{darlow2018cinic}, Tiny-ImageNet \cite{russakovsky2015imagenet} and ImageNet \cite{imagenet_cvpr09}. %
Dataset details and preprocessing methods are given in Appendix \ref{dataset_details}.

\begin{table}[t!]
\footnotesize
\centering
\caption{The classification accuracies of our model and the baselines. ``MN" denotes Meta-Neighborhoods. Results from three individual runs are reported and the best performance is marked as bold. Given that backbones like ResNet-56 are strong, our consistent improvement is notable.}
\begin{tabular}{llllll}
\hline
 Datasets & $\textit{vanilla}$ & MAXL &  \multicolumn{3}{c}{ours} \\
\cline{4-6}
 &   &   & $\textit{vanilla}$+iFiLM & MN & MN+iFiLM \\
\hline
\multicolumn{4}{l}{\textbf{Backbone:} 4-layer ConvNet} \\
\hline
MNIST    &  99.44$\pm$0.03\% &  99.60$\pm$0.02\% & 99.40$\pm$0.02\% &\textbf{99.62$\pm$0.03\%}  & 99.58$\pm$0.03\%      \\
SVHN   & 93.02$\pm$0.12\% & 94.06$\pm$0.10\% & 93.95$\pm$0.12\%&  94.46$\pm$0.09\% & $\textbf{94.92$\pm$0.09\%}$      \\
MNIST-M & 96.18$\pm$0.05\% & 96.85$\pm$0.06\% & 96.99$\pm$0.07\%  &96.55$\pm$0.04\% & $\textbf{97.40$\pm$0.05\%}$ \\
PACS & 92.55$\pm$0.08\% & 94.85$\pm$0.12\% &  94.45$\pm$0.09\% &95.19$\pm$0.10\% & $\textbf{95.22$\pm$0.09\%}$ \\
\hline
\multicolumn{4}{l}{\textbf{Backbone:} DenseNet40-BC} \\
\hline
CIFAR-10    & 94.53$\pm$0.10\% &  94.83$\pm$0.09\% & 94.87$\pm$0.08\%&  95.04$\pm$0.11\% & $\textbf{95.22$\pm$0.09\%}$        \\
CIFAR-100   & 73.92$\pm$0.12\% & 75.64$\pm$0.14\% & 74.66$\pm$0.13\% & 76.32$\pm$0.16\% & $\textbf{76.96$\pm$0.14\%}$    \\
CINIC-10    & 84.92$\pm$0.07\% & 85.42$\pm$0.07\%&    85.11$\pm$0.08\%&85.73$\pm$0.10\% &  $\textbf{86.02$\pm$0.07\%}$    \\
Tiny-Imagenet & 49.28$\pm$0.18\% & 50.94$\pm$0.16\%  &  50.86$\pm$0.14\% &53.27$\pm$0.18\% & $\textbf{54.36$\pm$0.15\%}$ \\
\hline
\multicolumn{4}{l}{\textbf{Backbone:} ResNet-29} \\
\hline
CIFAR-10    & 95.06$\pm$0.10\% & 95.31$\pm$0.09\% & 95.17$\pm$0.10\% & 95.56$\pm$0.09\% & $\textbf{95.58$\pm$0.10\%}$      \\
CIFAR-100    & 76.51$\pm$0.15\% &77.94$\pm$0.12\% & 77.16$\pm$0.14\% & 78.84$\pm$0.14\% & $\textbf{79.84$\pm$0.11\%}$    \\
CINIC-10     & 86.03$\pm$0.08\% & 86.34$\pm$0.06\%  & 86.64$\pm$0.06\% & 86.86$\pm$0.08\% & $\textbf{87.35$\pm$0.09\%}$\\
Tiny-Imagenet & 54.82$\pm$0.17\% & 56.29$\pm$0.14\% & 55.59$\pm$0.17\% & 57.36$\pm$0.15\% & $\textbf{57.94$\pm$0.14\%}$    \\
\hline
\multicolumn{4}{l}{\textbf{Backbone:} ResNet-56} \\
\hline
CIFAR-10    & 95.73$\pm$0.08\% & 96.06$\pm$0.07\%  &  96.08$\pm$0.08\%& 96.36$\pm$0.07\% & $\textbf{96.40$\pm$0.06\%}$      \\
CIFAR-100    &  79.64$\pm$0.13\% & 80.36$\pm$0.13\% & 80.04$\pm$0.12\% &80.58$\pm$0.10\%  & $\textbf{80.90$\pm$0.12\%}$     \\
CINIC-10     & 88.21$\pm$0.07\% & 88.30$\pm$0.05\% &  88.57$\pm$0.07\% &88.61$\pm$0.06\% & $\textbf{88.99$\pm$0.07\%}$     \\
Tiny-Imagenet& 57.92$\pm$0.12\% & 58.94$\pm$0.16\% &   58.31$\pm$0.15\%&60.05$\pm$0.12\% & $\textbf{60.78$\pm$0.13\%}$ \\
ImageNet& 48.41$\pm$0.14\% & 48.83$\pm$0.16\% &   52.03$\pm$0.12\%&51.85$\pm$0.12\% & $\textbf{54.23$\pm$0.13\%}$ \\
\hline
\end{tabular}
\label{table:main-result}
\end{table}

Our models are compared to two baselines: \textit{vanilla}, a traditional parametric ConvNet with the same architecture as ours but without the learnable dictionary, and MAXL \cite{liu2019self}, where an auxiliary label generator is meta-learned to enhance the primary classification tasks. 
For MNIST, MNIST-M, SVHN and PACS, a 4-layer ConvNet is selected as the feature extractor $\mu_\theta$. For the other four datasets,
three deep convolutional architectures, DenseNet40-BC \cite{huang2017densely}, ResNet29, and ResNet56 \cite{he2016identity}, are used as $\mu_{\theta}$. $f_{\phi}$ is implemented as a cos-similarity based classifier with one linear layer for both \textit{vanilla} and our models. Experiment details for our models and baselines are provided in Appendix \ref{classification_implementation}.
\paragraph{Results}
Table \ref{table:main-result} compares the test accuracy to baselines. We show that both Meta-Neighborhoods (MN) and iFiLM ($\textit{vanilla}$+iFiLM) can improve over $\textit{vanilla}$. The best performance is achieved when combining MN and iFiLM (MN+iFiLM), which outperforms the $\textit{vanilla}$ and MAXL baselines across several network architectures and different datasets. This indicates Meta-Neighborhoods and iFiLM are complementary and it is beneficial to adjust both $f_{\phi}$ and $\mu_{\theta}$ per-instance. Our method is also effective at PACS and MNIST-M datasets that contain significant domain shifts. 

Note that backbones like ResNet-56 are \emph{already powerful} for these datasets and there is limited room for improvement over the $\textit{vanilla}$ model. For instance, employing ResNet-110 instead of ResNet-56 \emph{only gives 0.14\% and 0.40\% further improvements} on CIFAR-10 and CIFAR-100, but at the expense of doubling the number of parameters. Yet Meta-Neighborhoods still consistently achieve greater improvements over $\textit{vanilla}$ than previous SOTA meta-learning method MAXL \cite{liu2019self}.
Compared to $\textit{vanilla}$ models, Meta-Neighborhoods with the same backbone architecture contains extra trainable parameters stored in the dictionary $M$. However, as discussed in Appendix \ref{vanilla_extra}, the performance boost in our paper originates from adjusting models using neighbors, rather than a naive increase in the number of parameters.

\begin{wrapfigure}{R}{0.55\textwidth}
\centering
\vspace{-25pt}
\includegraphics[width=0.55\textwidth]{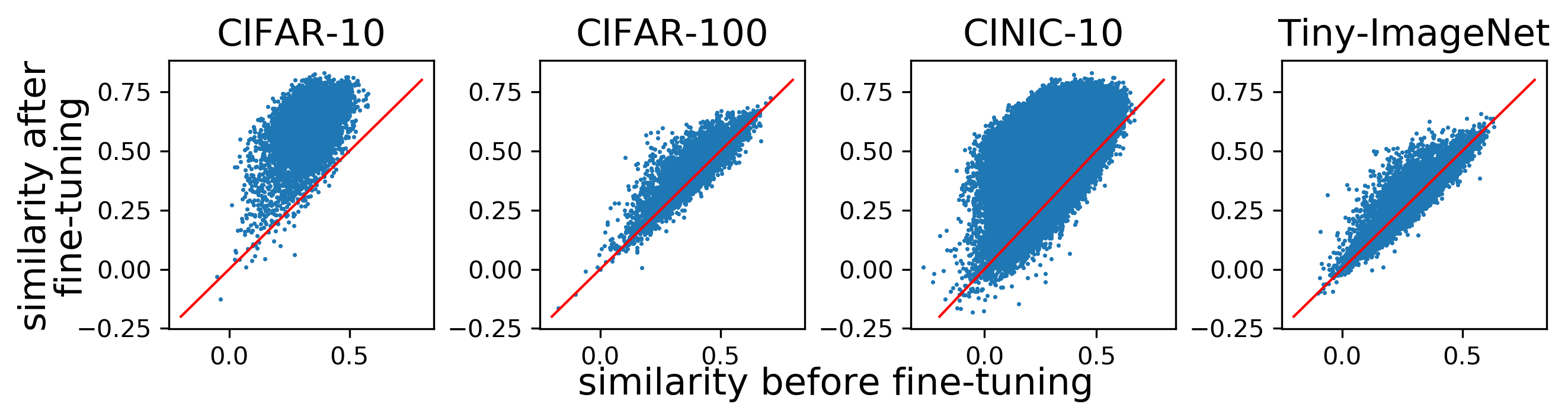} 
\caption{Cosine similarities between features and their corresponding ground-truth class prototypes. Each blue point denotes a testing sample. We expect most samples locate above the red lines, meaning larger similarities after finetuning.}
\vspace{-10pt}
\label{fig3}
\end{wrapfigure}

Since we implement $f_\phi$ as a cosine similarity classification layer, $\phi$ can be regarded as the prototypes for each class.
To verify that finetuning over neighborhoods helps with the classification, we compare the cosine similarity between the extracted feature $z_i$ and its corresponding ground-truth prototypes $\phi[y_i]$ before and after finetuning, where $y_i$ is the class label for $z_i$. From Fig.~\ref{fig3}, we can see that the cosine similarities increase after finetuning for most test examples, which indicates better predictions after finetuning.

We conduct ablation studies for $S$ and $T$ in Appendix \ref{ablation_study_S_gamma}. Ablations for the number of inner loop finetuning steps and different forms of $\alpha$ (scalar or diagonal) are provided in Appendix \ref{ablation_study_alpha}. Additional experiments for $\textit{vanilla}$ models trained with dot-product output layer and SGD are provided in Appendix \ref{ablation_vanilla}. We also discuss the inference speed of our model in Appendix \ref{inference_speed}. 

\begin{figure*}[ttt!]
\begin{tabular}{cc}
\begin{minipage}{.57\textwidth}
    \centering
    \subfigure[MNIST]{\includegraphics[width=0.49\linewidth]{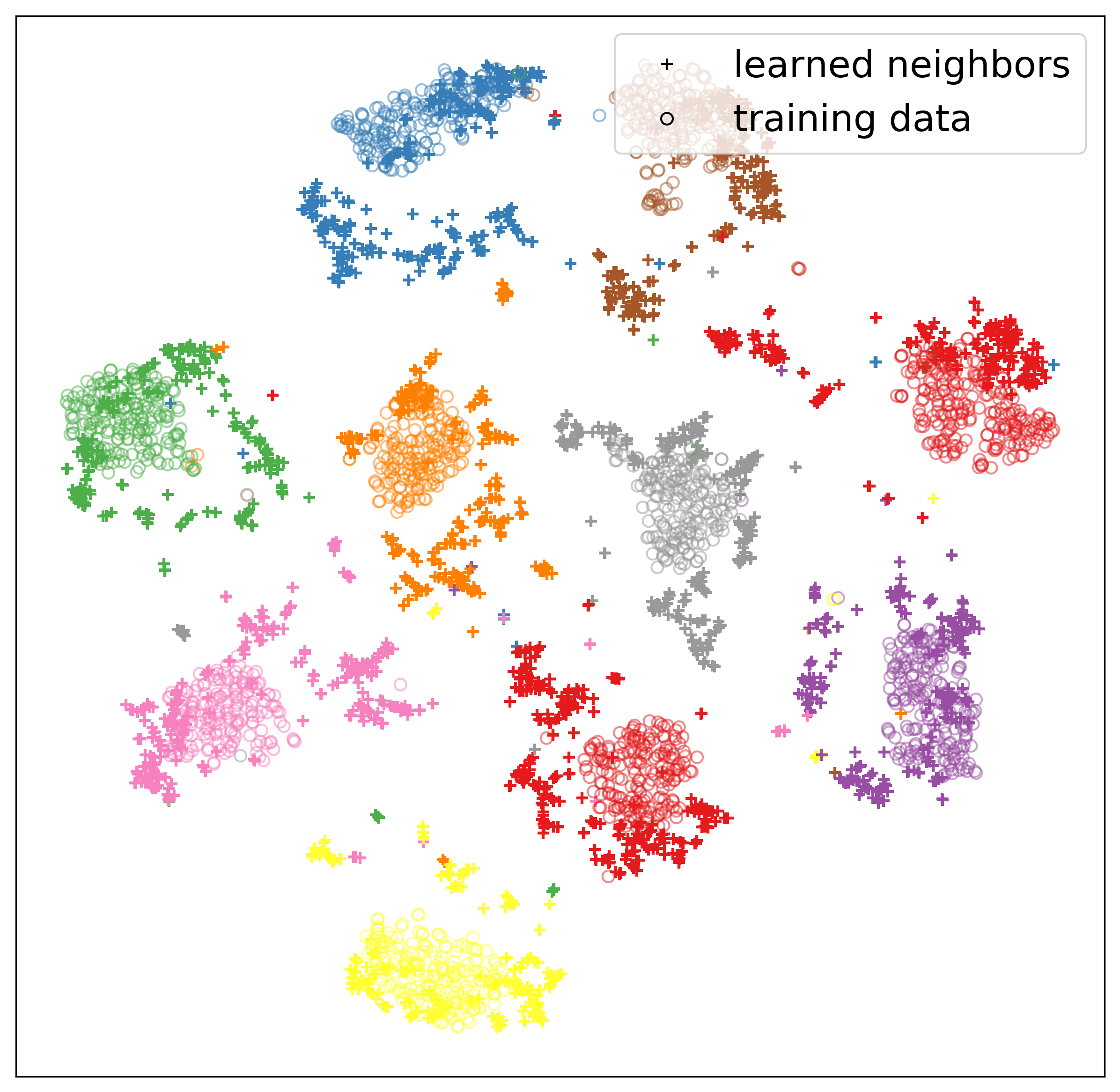}}
    \subfigure[CIFAR-10]{\includegraphics[width=0.49\linewidth]{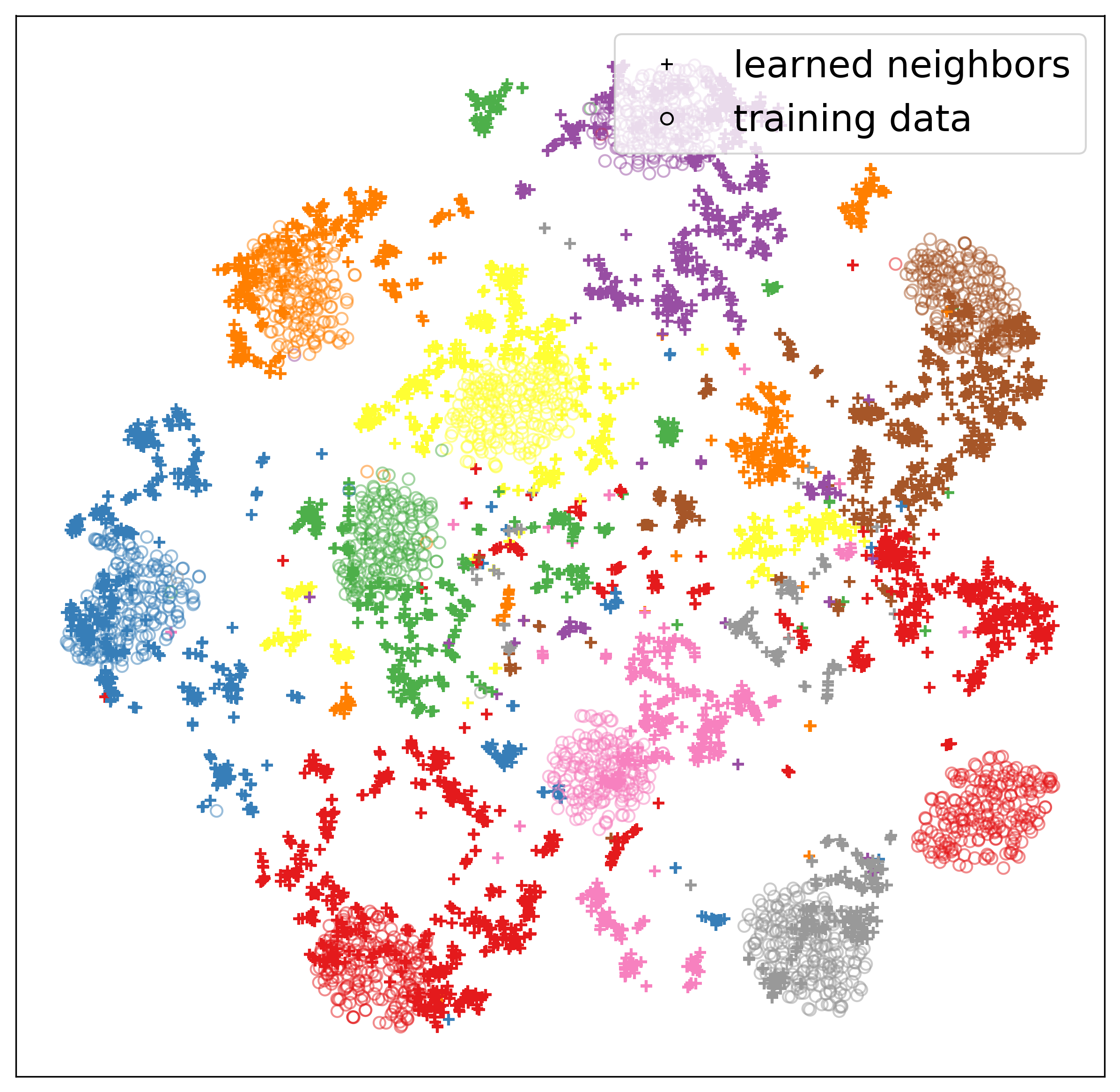}}
    \caption{t-SNE visualization of the learned neighbors and training data on MNIST and CIFAR-10. Learned neighbors are marked as ``+" and real training data are marked as ``o". The class information is represented by colors. Please zoom in to see the differences between ``+" and ``o".}
    \vspace{-10pt}
    \label{fig7}
\end{minipage} &
\hfill

    \begin{minipage}{.37\textwidth}
    \includegraphics[width=\columnwidth]{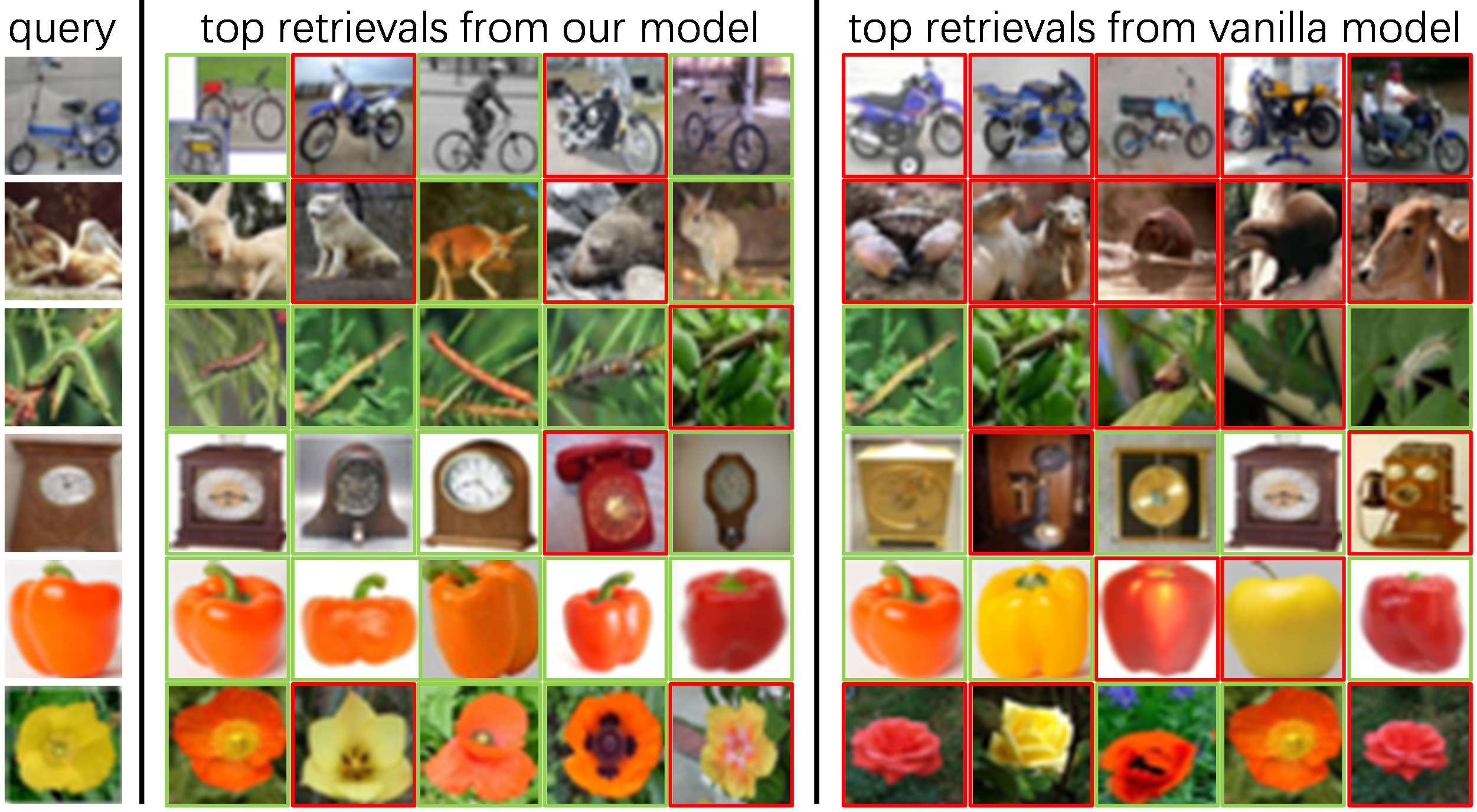} 
\caption{
Sub-category image retrieval quality of our model and the $\textit{vanilla}$ model. Correct retrievals have green outlines and wrong retrievals have red outlines.}
\label{fig6}
    \end{minipage}

\end{tabular}
\end{figure*}

\paragraph{Analysis of Learned Neighbors}
In Fig.~\ref{fig7}, we use t-SNE \cite{maaten2008visualizing} to visualize the 2D embeddings of the learned neighbors (marked as ``+") along with the training data (marked as ``o") on CIFAR-10 and MNIST. The 2D embeddings of training data are computed on the features $z_i=\mu_\theta(x_i)$, and embeddings of learned neighbors are computed on keys $k_j$. Classes of the learned neighbors are inferred from values $v_j$. It is shown that our model learns neighbors beyond the training set as the learned neighbors do not completely overlap with the training data, and the learned neighbors represent ``hard cases" around class boundaries to assist our model making better predictions. It is also interesting to note that this follows the same trend as our toy example in Fig.~\ref{fig1}.

We further investigate whether the learned neighbors are semantically meaningful by retrieving their 5-nearest neighbors from the test set. As shown in Appendix \ref{vis_neighbor_sub_category}, the retrieved 5-nearest neighbors for each learned neighbor not only come from the same class, but also represent a specific sub-category concept. In Appendix \ref{sub_category_discovery}, we quantitatively show that our method has superior sub-category discovery performance than $\textit{vanilla}$ on CIFAR-100: our method achieves 63.3\% accuracy on the 100 fine-grained classes while $\textit{vanilla}$ only achieves 59.28\%. This indicates our learned neighbors can preserve fine-grained details that are not explicitly given in the supervision signal. Qualitative results are shown in Fig.~\ref{fig6}.

\subsection{Regression}
\label{exp_regression}
\begin{wraptable}{r}{8cm}
\centering
\vspace{-50pt}
\tiny
\caption{Test MSE of our model, kNN and the $\textit{vanilla}$ baseline on five datasets. $n$ and $d$ respectively denote the dataset size and the data dimension.}
\setlength{\tabcolsep}{3pt}
\begin{tabular}{llllll}
\hline
Datasets & $n$ & $d$ & kNN & $\textit{vanilla}$    & Meta-Neighborhoods         \\ \hline
music & 515345 &  90 & 0.6812$\pm$0.0062    & 0.6236$\pm$0.0056     & $\textbf{0.6088$\pm$0.0050}$     \\
toms & 28179 &  96 &  0.0602$\pm$0.0083   & 0.0594$\pm$0.0080     & $\textbf{0.0531$\pm$0.0073}$     \\
cte &  53500 & 384  & 0.00134$\pm$0.00023 & 0.00121$\pm$0.00022 & $\textbf{0.00109$\pm$0.00015}$ \\
super & 21263 & 80  & 0.1126$\pm$0.0061 &  0.1132$\pm$0.0060     & $\textbf{0.1077$\pm$0.0068}$     \\
gom  & 1059 & 116      & 0.5982$\pm$0.0521 & 0.5949$\pm$0.0515     & $\textbf{0.5681$\pm$0.0563}$     \\ \hline
\end{tabular}
\vspace{-5pt}
\label{result:regression}
\end{wraptable}
We use five publicly available datasets with various sizes from UCI Machine Learning Repository \cite{Dua:2019}.
For regression tasks, we found learning neighbors in the input space yields better performance compared to learning neighbors in the feature space. As a result, the feature extractor $\mu_{\theta}$ is implemented as an identity mapping function.
We compare our model to kNN and $\textit{vanilla}$ baseline using the mean square error (MSE). The $\textit{vanilla}$ baseline is a multilayer perceptron for regression. We searched for the best network configuration for the $\textit{vanilla}$ model on every dataset by varying the number of layers in $\{2,3,4,5\}$ and the number of neurons at each layer in $\{32,64,128,256\}$. For each dataset, our model uses the same network architecture to $\textit{vanilla}$. Model details, training details, and hyperparameter settings are given in Appendix \ref{regression_implementation}. 5-fold cross-validation is used to report the results in Table \ref{result:regression}. Our model has lower MSE scores compared to the $\textit{vanilla}$ model across the five datasets. The results of Meta-Neighborhoods and $\textit{vanilla}$ are statistically different based on paired Student's t-test with a significance of 0.05. We found that naively increasing the model complexity for $\textit{vanilla}$ baseline can not further improve its performance due to over-fitting, but our method can as it takes advantage of non-parametric neighbor information.

\section{Conclusion}
In this work, we introduced Meta-Neighborhoods, a novel meta-learning framework that adjusts predictions based on learnable neighbors.
It is interesting to note that in addition to directly generalizing KNN, Meta-Neighborhoods provides a learning paradigm that aligns more closely with human learning.
Human learning jointly leverages previous examples both to shape the perceptual features we focus on and to pull relevant memories when faced with novel scenarios \cite{kuhl2003foreign}.
In much the same way, Meta-Neighborhoods use feature-based models that are then adjusted by pulling memories from previous data.
We show through extensive empirical studies that Meta-Neighborhoods improve the performance of already strong backbone networks like DenseNet and ResNet on several benchmark datasets.
In addition to providing a greater gain in performance than previous state-of-the-art meta-learning methods like MAXL, Meta-Neighborhoods also works both for regression and classification, and provides further interpretability.

\nocite{langley00}

\section*{Broader Impact}

Any general discriminative machine learning model runs the risk of making biased and offensive predictions reflective of training data. Our work is no exception as it aims at improving discriminative learning performance. To reduce these negative influences to the minimum possible extent, we only use standard benchmarks in this work, such as CIFAR-10, Tiny-ImageNet, MNIST, and datasets from the UCI machine learning repository.

Our work does impose some privacy concerns as we are learning a per-instance adjusted model in this work. Potential applications of the proposed model include precision medicine, personalized recommendation systems, and personalized driver assistance systems. To keep user data safe, it is desirable to only deploy our model locally.

The induced neighbors in our work, which are semantically meaningful, can also be regarded as fake synthetic data. Like DeepFakes, they may also raise a set of challenging policy, technology, and legal issues. Legislation regarding synthetic data should take effect and the research community needs to develop effective methods to detect these synthetic data.

\bibliography{neurips_2020}
\bibliographystyle{plainnat.bst}

\clearpage

\appendix

\section{Experiment Details for Classification} 

\subsection{Dataset Details}
\label{dataset_details}
All datasets except ImageNet were resized to resolution 32$\times$32 to facilitate fast experimentation. ImageNet was resized to resolution 64$\times$64.
CIFAR-10/100 are image classification datasets containing a training set of 50K and a testing set of 10K 32$\times$32 color
images across the 10/100 classes. CINIC-10 has 270K 32$\times$32 images across 10 classes equally split into three parts for training, validation, and testing. Tiny-ImageNet has a training set of 100K and a testing set of 10K 64$\times$64 images across the 200 classes. PACS contains domain shifts (Art painting, Cartoon, Photo and Sketch), and we train on all domains to test our model's ability to fit on all domains. PACS is split as 9K/1K/10K for training/validation/testing. MNIST-M also contains domain shifts due to the random background. MNIST-M is split as 55K/5K/10K for training/validation/testing.

\subsection{Implementation Details}
\label{classification_implementation}
We find adaptive learning rate optimizers like Adam \cite{kingma2014adam} are more effective than SGD at optimizing the dictionary, as these optimizers can compute individual adaptive learning rates for different parameters. In all the experiments, we adopt AdamW \cite{loshchilov2018decoupled}, a variant of Adam \cite{kingma2014adam} with correct weight decay mechanism and better generalization capability. 

Our models are trained by AdamW with weight decay rate 7.5e-5, an initial learning rate of 1e-3 and batch size 128. For CIFAR-10 and CIFAR-100, we train for 400 epochs with learning rate reduced to 1e-4 at epoch 300. For CINIC-10 and Tiny-ImageNet, models are trained for 350 epochs with the learning rate reduced to 1e-4 at epoch 250. We train $\textit{vanilla}$ baselines by AdamW in the same way as our models.

We initialize $k_{j}$ and $v_{j}$ with Gaussian distribution $\mathcal{N}(0,\,0.01)$, and apply a softmax over $v_{j}$ to make it a probability distribution over multiple classes. Cosine similarity is adopted as the similarity metric in \eqref{eq:sim}. We set $T$ in \eqref{eq:sim} to 0.2. For CIFAR-10 and CINIC-10, the number of dictionary entries $S$ is set to 5000, and for CIFAR-100 and Tiny-Imagenet $S$ is set to 10000 in the main results. For the results in Table \ref{table:main-result}, we set the number of inner loop steps to 1 and set $\alpha$ to be a learnable scalar. Ablation studies of the number of inner loop finetuning steps and different forms of $\alpha$ (scalar or diagonal) are given in Appendix \ref{ablation_study_alpha}.

When optimizing the dictionary at the early stage, some entries might always receive larger gradients than others due to the random initialization, which causes the model exploiting only part of the dictionary entries. To boost the diversity of used entries, we randomly dropout 50\% of the entries during training. This also helps to prevent overfitting and improve training speed.

For models (Densenet40-BC, ResNet) with iFiLM, we learn a learnable fix-sized dictionary $M_{l}=\{k^l_j,\gamma^l_j, \beta^l_j\}_{j=1}^{S_l}$ with every batch normalization layer, where $S_l$ is set to 10 for all $M_l$. 

We ran MAXL using code provided in \cite{liu2019self} and performed extensive hyperparameter tuning to report the best results. We run MAXL with SGD of a learning rate of 0.1 with momentum 0.9 and weight decay $5\times10^{−4}$, and hierarchies to set to 5 as recommended in MAXL. We also try a larger learning rate of 0.01. We dropped the learning rate by half for every 50 epochs with a total of 200 epochs.

\begin{figure}[b!]
\center
\subfigure[ResNet-29]{\includegraphics[width=0.32\linewidth]{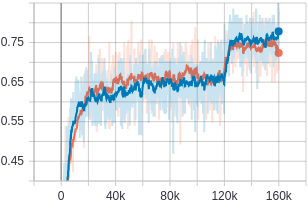}}
\subfigure[ResNet-56]{\includegraphics[width=0.32\linewidth]{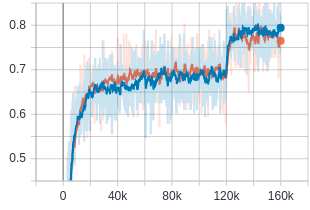}}
\subfigure[DenseNet40-BC]{\includegraphics[width=0.32\linewidth]{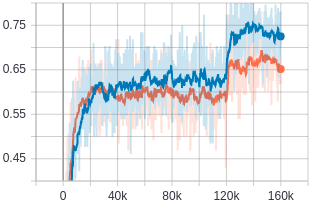}}

  \caption{Validation accuracies of $\textit{vanilla}$ models with (shown in orange) and without (shown in blue) extra parameters on CIFAR-100 across three backbone architectures.}
  \label{fig8}
\end{figure}

\subsection{$\textit{vanilla}$ Models with Extra Parameters} 
\label{vanilla_extra}
Compared to $\textit{vanilla}$ models, our best model (MN+iFiLM) with the same backbone architectures contain $S\times(d+C)$ extra trainable parameters in the dictionary $M$ and $3\times S^l \times C^{l}$ extra trainable parameters in $M^l$, where $S$ is the number of dictionary entries in $M$, $d$ is the dimension of the feature vector, $C$ is the number of classes, $S^l$ is the number of dictionary entries in $M^l$ and $C^l$ is the number of feature map channels. Since we set a small $S^l=10$, extra parameters in $M^l$ is negligible compared to the remaining model parameters. Therefore we only consider the extra parameters in $M$ in this section.

To investigate whether $\textit{vanilla}$ models can also benefit from more parameters as our model, we add an extra fully connected layer with the same number of extra parameters to $\textit{vanilla}$ models. According to Fig.~\ref{fig8}, $\textit{vanilla}$ models with extra parameters have inferior validation accuracy than $\textit{vanilla}$ models without extra parameters across three backbone architectures on CIFAR-100. The same observation holds for other datasets. Therefore, $\textit{vanilla}$ models can not benefit from more parameters as our model, and the performance boost in our paper originates from finetuning using neighbors, rather than a naive increase in parameters.

\begin{figure}[b]
\centering
\includegraphics[width=0.8\columnwidth]{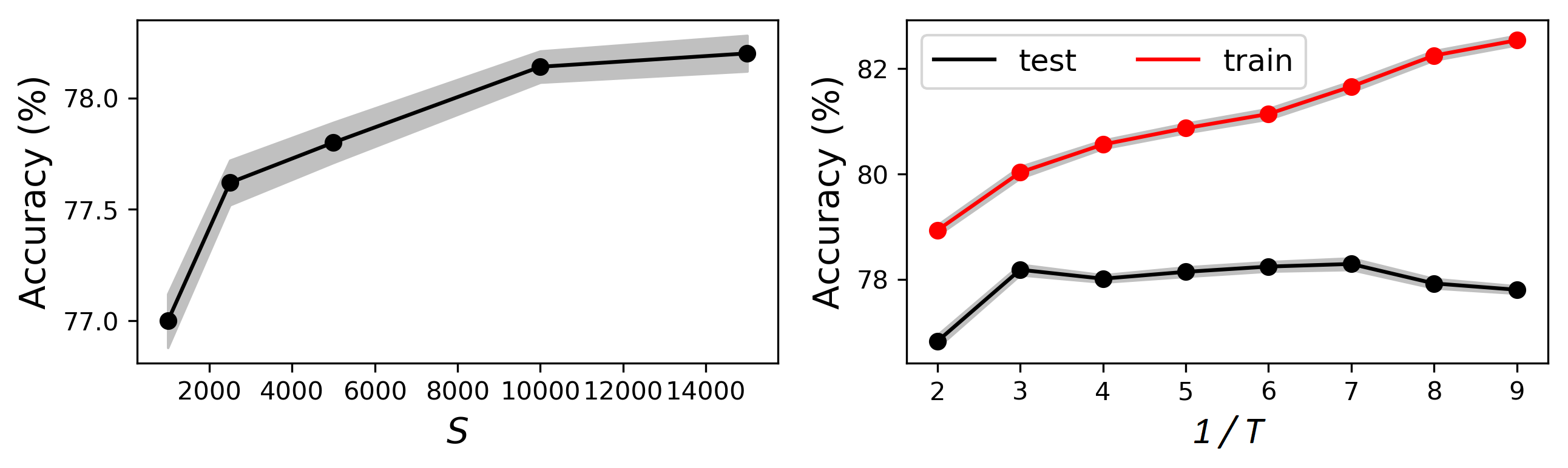} 
\caption{Ablation studies of $S$ and $T$ on CIFAR-100. 
}
\label{fig4}
\end{figure}

We show in Table \ref{table:main-result} that our method consistently improves performance regardless of the choice of backbones, indicating our method can further improve over an even bigger model.

Note that our method is particularly helpful for low-capacity models, which usually handle simpler tasks like MNIST/SVHN classification and regression. In the regression experiment in Section \ref{exp_regression}, we show that naively increasing the model capacity can’t effectively further improve performance on these simple tasks due to over-fitting, but our method can as it takes advantage of non-parametric neighbor information.

\subsection{Ablation Study of $S$ and $T$}
\label{ablation_study_S_gamma}
We investigate the impact of hyperparameters, $S$ and $T$ on CIFAR-100 using ResNet-29. As shown in Fig.~\ref{fig4}, the testing accuracy increases with the increase of the number of dictionary entries $S$ (with $T$ set to 0.2), which indicates a better finetuning of $\phi$ with the help of more learnable neighbors. The temperature $T$ controls the ``peakiness" of the similarity distribution in \eqref{eq:sim}. It is set to a fixed value rather than learned in all experiments. If we enable $T$ to be learnable, it always grows to a small value, which makes the model only pay attention to a small number of entries and leads to over-fitting. On the other hand, if $T$ is too large, the model will pay uniform attention to all entries in $M$, which leads to under-fitting. According to Fig.~\ref{fig4}, the optimal range of $\frac{1}{T}$ is [3,7] (with $S$ set to 10000).

\subsection{Ablation Study of the Number of Inner Loop Steps and the Form of $\alpha$}
\label{ablation_study_alpha}
As shown in Table \ref{table:ablation-result}, we found that implementing the inner loop learning rate $\alpha$ as a learnable scalar usually gives better performance. The majority of our results are better when setting the number inner-loop finetuning steps to 1. As we learn $\alpha$ rather than set it to a fixed value, our model can achieve good performance in only one step of finetuning. 

We also investigate how Meta-Neighborhoods performs without finetuning at test time. With DenseNet40, our pre-tuned and post-tuned models achieve 90\%/95\% on cifar10 and 69\%/76\% on cifar100, indicating it is beneficial to finetune the model. This is also substantiated by Fig.~\ref{fig3}. This bad result without finetuning at the testing stage is unsurprising because the model is used differently at the training stage and the testing stage.
\subsection{Ablation Study of $\textit{vanilla}$ Models}
\label{ablation_vanilla}
To ensure that the modifications (using cos-similarity output layer and AdamW optimizer) that we made to the commonly-used $\textit{vanilla}$ model trained by SGD with dot-product output layer do not deteriorate the performance, we also report the accuracies of $\textit{vanilla}$ models with either dot-product or cos-similarity output layer and trained either by AdamW or SGD in Table \ref{table:ablation-result}. All these four $\textit{vanilla}$ baselines have similar performance that is worse compared to our method.

\begin{table}[t]
\scriptsize
\centering
\caption{The classification accuracies of our model and the baselines. ``ft" in our methods denotes how many finetuning steps are used in the inner loop. ``S" in our methods denotes using a scalar inner loop learning rate, while ``D" denotes using a diagonal matrix inner loop learning rate.}
\begin{tabular}{llllllllll}
\hline
 & \multicolumn{4}{c}{$\textit{vanilla}$} & MAXL &  \multicolumn{4}{c}{Meta-Neighborhoods (ours)} \\
\cline{2-5}\cline{7-10}
Datasets &dot-sgd & dot-adamw  & cos-sgd  & cos-adamw  &   & ft:1+S & ft:1+D & ft:3+S & ft:3+D \\
\hline
\multicolumn{4}{l}{\textbf{Backbone:} 4-layer ConvNet} \\
\hline
MNIST    & 99.39\% &  99.47\%   & 99.42\% & 99.44\% &  99.60\% & \textbf{99.62\%} & 99.45\% & 99.50\% & 99.55\%      \\
SVHN   &  93.01\%  & 93.12\%    & 92.93\% & 93.02\% & 94.06\% & $\textbf{94.46\%}$ & 93.95\% & 94.05\% & 94.05\%      \\

\hline
\multicolumn{4}{l}{\textbf{Backbone:} DenseNet40-BC} \\
\hline
CIFAR-10    & 94.56\% &  94.46\%   & 94.52\% & 94.53\% &  94.83\% & 95.04\% & 94.79\% & 95.08\% & $\textbf{95.12\%}$       \\
CIFAR-100   &  73.85\%  & 73.68\%    & 74.08\% & 73.92\% & 75.64\% & 76.32\% & $\textbf{77.20\%}$ & 76.04\% & 76.42\%      \\
CINIC-10    &  85.13\% &  85.02\%   & 85.10\% & 84.92\% & 85.42\%&  85.73\% & $\textbf{85.76\%}$ & 85.51\% & 85.21\%      \\
Tiny-Imagenet(32$\times$32) & 49.32\% &  49.21\%   & 49.40\% & 49.28\% & 50.94\%  & $\textbf{53.27\%}$ & 53.16\% & 52.88\% & 52.61\%      \\
\hline
\multicolumn{4}{l}{\textbf{Backbone:} ResNet-29} \\
\hline
CIFAR-10    & 94.91\%  &  94.96\%   & 95.02\% & 95.06\% & 95.31\% &$\textbf{95.56\%}$ &  95.28\% & 95.36\% & 95.26\%      \\
CIFAR-100    &  76.65\%  &  76.72\%   & 76.70\% & 76.51\% &77.94\% & $\textbf{78.84\%}$ & 78.20\% & 78.04\% & 78.40\%      \\
CINIC-10     &  85.86\% &   85.91\%  & 85.96\% & 86.03\% & 86.34\%  & $\textbf{86.86\%}$ & 86.41\% & 86.38\% & 86.51\%      \\
Tiny-Imagenet(32$\times$32) &54.79\% &  54.67\%   & 54.97\% & 54.82\% & 56.29\% & 57.36\% & 56.93\% & $\textbf{57.64\%}$ & 57.27\%      \\
\hline
\multicolumn{4}{l}{\textbf{Backbone:} ResNet-56} \\
\hline
CIFAR-10    & 95.64\%  &    95.83\% & 95.71\% & 95.73\% & 96.06\%  & $\textbf{96.36\%}$ & 96.32\% & 96.28\% & 96.04\%      \\
CIFAR-100    &  79.54\%  & 79.68\%    & 79.78\% & 79.64\% & 80.36\% & 80.58\% & \textbf{80.66}\% & 80.20\% & 80.14\%      \\
CINIC-10     & 88.03\%  &  88.15\%   & 87.90\% & 88.21\% & 88.30\% & $\textbf{88.61\%}$ & 88.47\% & 88.42\% & 88.38\%      \\
Tiny-Imagenet(32$\times$32)& 57.79\% &  57.95\%   & 57.89\% & 57.92\% & 58.94\% & $\textbf{60.05\%}$ & 59.20\% & 59.85\% & 59.88\%      \\
\hline
\end{tabular}
\label{table:ablation-result}
\end{table}

\subsection{Inference Speed}
\label{inference_speed}
Due to the neighbor searching process and finetuning process, our method is slower at the testing time compared to the $\textit{vanilla}$ testing process which only requires a single feed-forward propagation. However, our method is only approximately 2 times slower than the vanilla models due to the following reasons: (1) our searching space is small (only 5000 neighboring points) (2) the attention calculation and the finetuning step are parallelized efficiently across multiple GPU threads (3) for the classification task, we only finetune the parameters of the classification layer rather than the whole model (4) we only finetune for a small number of steps (1 or 3).

\subsection{Visualizing Learned Neighbors by Retrieving Real Neighbors}
\label{vis_neighbor_sub_category}
We further investigate whether the learned neighbors are semantically meaningful by retrieving their 5-nearest neighbors from the test set.

\begin{figure}[t]
\centering
\includegraphics[width=0.8\columnwidth]{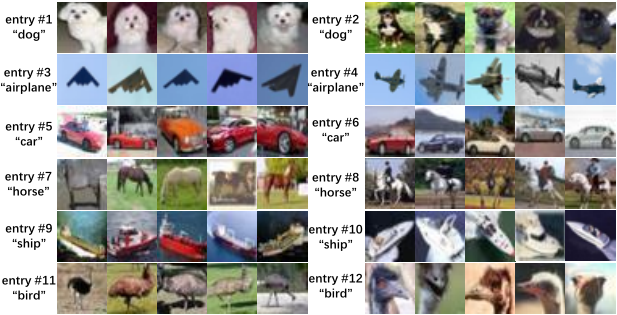} 
\caption{
5-nearest neighbors of 12 dictionary entries retrieved using $k_j$ in \eqref{inner_loop_feature_space} from the CIFAR-10 test set. Entry indexes and entry classes inferred from $v_j$ are shown on the left of each group of images. By comparing the two entries on the same row, we discover that different entries represent different fine-grained sub-category concepts.  
}
\label{fig5}
\end{figure}

\begin{figure}[]
\centering
\includegraphics[width=0.8\columnwidth]{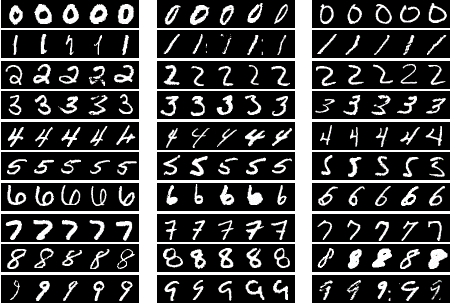} 
\caption{5-nearest neighbors of 30 dictionary entries retrieved using $k_j$ in  \eqref{inner_loop_feature_space} from the MNIST test set. By comparing the three entries on the same row, we discover that different entries represent different fine-grained attributes such as stroke widths, character orientations and fonts.  
}
\label{mnist_vis}
\end{figure}

Examples on CIFAR-10 and MNIST are respectively shown in Fig.~\ref{fig5} and \ref{mnist_vis}. We found most entries can retrieve consistent neighbors. It is shown that the retrieved 5-nearest neighbors for each learned neighbor not only come from the same class, but also represent a specific sub-category concept. For instance, both of the entries on the fifth row of Fig.~\ref{fig5} represent ``ship", but the first represents ``steamship" while the second represents ``speedboat".

\subsection{Sub-category Discovery}
\label{sub_category_discovery}
To quantitatively measure the sub-category discovery performance, we train our model ($S$ is set to 5000) and the $\textit{vanilla}$ cos-adamw model with the same Densenet40 backbone on CIFAR-100 only using its coarse-grained annotations (20 classes), and evaluate the classification accuracy on the fine-grained 100 categories using KNN classifiers, which is called induction accuracy as in \cite{huh2016makes}. For the $\textit{vanilla}$ model, the KNN classifier uses the feature $z_i=\mu_\theta(x_i)$, while for our model, the classifier uses the attention vector $\vec{\omega}=[\omega(z_i,k_1),\omega(z_i,k_2),...,\omega(z_i,k_S)]$ over all entries in $M$. Our KNN and $\textit{vanilla}$'s KNN achieve similar accuracy on the coarse 20 classes (80.18\% versus 80.30\%). However, on the fine-grained 100 classes, our KNN achieves 63.3\% while the $\textit{vanilla}$'s KNN only achieves 59.28\%. This indicates our learned neighbors can preserve fine-grained details that are not explicitly given in the supervision signal. Examples of nearest neighbors retrieved are shown in Fig.~\ref{fig6}.
\section{Experiment Details for Regression}
\label{regression_implementation}
We use five publicly available datasets with various sizes from UCI Machine Learning Repository: \textit{music} (YearPredictionMSD), \textit{toms}, \textit{cte} (Relative location of CT slices on axial axis), \textit{super} (Superconduct), and \textit{gom} (Geographical Original of Music). All datasets are normalized dimension-wise to have zero means and unit variances.

For regression tasks, we found learning neighbors in the input space yields better performance compared to learning neighbors in the feature space. As a result, our model for regression only consists of an output network $f_\phi$ and a dictionary $M$. It is trained with the loss in \eqref{eq:outer-loss}.
A learning rate of 1e-3 and a batch size of 128 are used, and the best weight decay rate is chosen for each dataset. The training stops if the validation loss does not reduce for 10 epochs. We initialize $k_{j}$ with Gaussian distribution $\mathcal{N}(0,\,0.01)$ and $v_{j}$ with uniform distribution in the same range of the regression labels. Cosine similarity is adopted to implement the similarity metric in \eqref{eq:sim}. We use 1000 dictionary entries and set $T$ to 0.1 based on the validation performance. Because there is no batch normalization layer in $f_{\phi}$, iFiLM is not used in this regression experiment.

\end{document}